\documentclass[sigconf]{acmart}

\usepackage{soul}
\usepackage{url}
\usepackage{hyperref}
\usepackage[utf8]{inputenc}
\usepackage{caption}
\usepackage{xcolor}
\usepackage{graphicx}
\usepackage{caption}
\usepackage{courier}
\usepackage{multirow}
\usepackage{subcaption}
\usepackage{color}
\usepackage{todonotes}
\usepackage{amssymb}
\usepackage{gensymb}
\usepackage{enumitem}
\usepackage[linesnumbered, ruled]{algorithm2e}
\usepackage{amsmath}

\copyrightyear{2020}
\acmYear{2020}
\setcopyright{acmlicensed}
\acmConference[WSDM '20]{The Thirteenth ACM International Conference on Web Search and Data Mining}{February 3--7, 2020}{Houston, TX, USA}
\acmBooktitle{The Thirteenth ACM International Conference on Web Search and Data Mining (WSDM '20), February 3--7, 2020, Houston, TX, USA}
\acmPrice{15.00}
\acmDOI{10.1145/3336191.3371851}
\acmISBN{978-1-4503-6822-3/20/02}
\newtheorem{mypro}{Problem}
\newcommand{\our}{{PA-GNN}\xspace}
\newcommand{\ours}{{PA-GNN}\xspace}

\def \E {\mathcal{E}}
\def \G {\mathcal{G}}
\def \V {\mathcal{V}}
\def \N {\mathcal{N}}
\def \P {\mathcal{P}}
\def \L {\mathcal{L}}
\def \T {\mathcal{T}}

\def \R {\mathbb{R}}

\def \bg {\mathbf{G}}

\def \X {\mathbf{X}}

\def \H {\mathbf{H}}
\def \h {\mathbf{h}}
\def \W {\mathbf{W}}
\def \S {\mathbf{S}}
\def \Q {\mathbf{Q}}
\def \h {\mathbf{h}}
\def \x {\mathbf{x}}

\title{Transferring Robustness for Graph Neural Network Against Poisoning Attacks}

\author{Xianfeng Tang$^\dagger$, Yandong Li$^{\ddagger}$, Yiwei Sun$^\dagger$, Huaxiu Yao$^\dagger$, Prasenjit Mitra$^\dagger$, Suhang Wang$^\dagger$}\authornote{Corresponding Author.}
\affiliation{
\institution{Pennsylvania State University$^\dagger$, University of Central Florida$^\ddagger$}
}
\affiliation{
  \institution{\{tangxianfeng, lyndon.leeseu\}@outlook.com  \{yus162, huy144, pum10, szw494\}@psu.edu}
}
\renewcommand{\shortauthors}{X. Tang et al.}

\begin{document}
\fancyhead{}
\begin{abstract}
Graph neural networks (GNNs) are widely used in many applications. 
However, their robustness against adversarial attacks is criticized. 
Prior studies show that using unnoticeable modifications on graph topology or nodal features can significantly reduce the performances of GNNs.
It is very challenging to design robust graph neural networks against poisoning attack and several efforts have been taken. 
Existing work aims at reducing the negative impact from adversarial edges only with the poisoned graph, which is sub-optimal since they fail to discriminate adversarial edges  from normal ones.
On the other hand, clean graphs from similar domains as the target poisoned graph are usually available in the real world. 
By perturbing these clean graphs, we create supervised knowledge to train the ability to detect adversarial edges so that the robustness of GNNs is elevated.
However, such potential for clean graphs is neglected by existing work.
To this end, we investigate a novel problem of improving the robustness of GNNs against poisoning attacks by exploring clean graphs.
Specifically, we propose \ours, which relies on a penalized aggregation mechanism that directly restrict the negative impact of adversarial edges by assigning them lower attention coefficients.
To optimize \ours for a poisoned graph, we design a meta-optimization algorithm that trains \ours to penalize perturbations using clean graphs and their adversarial counterparts, and transfers such ability to improve the robustness of \ours on the poisoned graph.
Experimental results on four real-world datasets demonstrate the robustness of \our against poisoning attacks on graphs.
Code and data are available here: \textcolor{blue}{\url{https://github.com/tangxianfeng/PA-GNN}}.
\end{abstract}

% <concept_id>10010147.10010257.10010258.10010261.10010276</concept_id>
% <concept_desc>Computing methodologies~Adversarial learning</concept_desc>
% <concept_id>10010147.10010257.10010282.10011305</concept_id>
% <concept_desc>Computing methodologies~Semi-supervised learning settings</concept_desc>
% <concept_id>10010147.10010257.10010293.10010294</concept_id>
% <concept_desc>Computing methodologies~Neural networks</concept_desc>

% \ccsdesc[500]{Computing methodologies~Adversarial learning}
% \ccsdesc[500]{Computing methodologies~Semi-supervised learning settings}
% \ccsdesc[500]{Computing methodologies~Neural networks}

\keywords{Robust Graph Neural Networks, Adversarial Defense}

\maketitle

\section{Introduction}
Graph neural networks (GNNs) \cite{kipf2016semi, hamilton2017inductive}, which explore the power of neural networks for graph data, have achieved remarkable results in various applications \cite{velivckovic2017graph,fan2019graph,huang2019online}.
The key to the success of GNNs is its signal-passing process \cite{wu2019simplifying}, where information from neighbors is aggregated for every node in each layer. The collected information enriches node representations, preserving both nodal feature characteristics and topological structure.

Though GNNs are effective for modeling graph data, the way that GNNs aggregate neighbor nodes' information for representation learning makes them vulnerable to adversarial attacks~\cite{zugner2018adversarial, zugner2018adversarial2, dai2018adversarial, wu2019adversarial, xu2019topology}.
Poisoning attack on a graph~\cite{zugner2018adversarial}, which adds/deletes carefully chosen edges to the graph topology or injects carefully designed perturbations to nodal features, can contaminate the neighborhoods of nodes, bring noises/errors to node representations, and degrade the performances of GNNs significantly. 
The lack of robustness become a critical issue of GNNs in many applications such as financial system and risk management \cite{akoglu2015graph}. For example, fake accounts created by a hacker can add friends with normal users on social networks to promote their scores predicted by a GNN model. A model that's not robust enough to resist such ``cheap'' attacks could lead to serious consequences.
Hence, it is important to develop robust GNNs against adversarial attacks.
Recent studies of adversarial attacks on GNNs suggest that adding perturbed edges is more effective than deleting edges or adding noises to node features~\cite{wu2019adversarial}. This is because node features are usually high-dimensional, requiring larger budgets to attack. Deleting edges only result in the loss of some information while adding edges is cheap to contaminate information passing dramatically. For example, adding a few bridge edges connecting two communities can affect the latent representations of many nodes. Thus, we focus on \textit{defense against the more effective poisoning attacks that a training graph is poisoned with injected adversarial edges}.

To defend against the injected adversarial edges, a natural idea is to delete these adversarial edges or reduce their negative impacts. Several efforts have been made in this direction~\cite{zhu2019robust,wu2019adversarial,jin2019power}. For example, \citeauthor{wu2019adversarial} \cite{wu2019adversarial} utilize Jaccard similarity of features to prune perturbed graphs with the assumption that connected nodes have high feature similarity. RGCN in~\cite{zhu2019robust} introduce Gaussian constrains on model parameters to absorb the effects of adversarial changes.
The aforementioned models only rely on the poisoned graph for training, leading to sub-optimal solutions.
The lack of supervised information about real perturbations in a poisoned graph obstructs models from modeling the distribution of adversarial edges.
Therefore, exploring alternative supervision for learning the ability to reduce the negative effects of adversarial edges is promising.

There usually exist clean graphs with similar topological distributions and attribute features to the poisoned graph.
For example, Yelp and Foursquare have similar co-review networks where the nodes are restaurants and two restaurants are linked if the number of co-reviewers exceeds a threshold. 
Facebook and Twitter can be treated as social networks that share similar domains. It is not difficult to acquire similar graphs for the targeted perturbed one. As shown in existing work~\cite{shu2018crossfire,lee2017transfer}, because of the similarity of topological and attribute features, we can transfer knowledge from source graphs to target ones so that the performance on target graphs is elevated.
Similarly, we can inject adversarial edges to clean graphs as supervisions for training robust GNNs, which are able to penalize adversarial edges. Such ability can be further transferred to improve the robustness of GNNs on the poisoned graph. 
Leveraging clean graphs to build robust GNNs is a promising direction. However, prior studies in this direction are rather limited.

Therefore, in this paper, we investigate a novel problem of exploring clean graphs for improving the robustness of GNNs against poisoning attacks.
The basic idea is first learning to discriminate adversarial edges, thereby reducing their negative effects, then transferring such ability to a GNN on the poisoned graph.
In essence, we are faced with two challenges: (\textbf{i}) how to mathematically utilize clean graphs to equip GNNs with the ability of reducing negative impacts of adversarial edges; and (\textbf{ii}) how to effectively transfer such ability learned on clean graphs to a poisoned graph.
In an attempt to solve these challenges, we propose a novel framework \underline{P}enalized \underline{A}ggregation \underline{GNN} (\our). Firstly, clean graphs are attacked by adding adversarial edges, which serve as supervisions of known perturbations. With these known adversarial edges, a penalized aggregation mechanism is then designed to learn the ability of alleviating negative influences from perturbations. We further transfer this ability to the target poisoned graph with a special meta-optimization approach, so that the robustness of GNNs is elevated. To the best of our knowledge, we are the first one to propose a GNN that can directly penalize perturbations and to leverage transfer learning for enhancing the robustness of GNN models.
The main contributions of this paper are:
\begin{itemize}[leftmargin=*]
    \item We study a new problem and propose a principle approach of exploring clean graphs for learning a robust GNN against poisoning attacks on a target graph;
    \item We provide a novel framework \our, which is able to alleviate the negative effects of adversarial edges with carefully designed penalized aggregation mechanism, and transfer the alleviation ability to the target poisoned graph with meta-optimization;
    \item We conduct extensive experiments on  real-world datasets to demonstrate the effectiveness of \our against various poisoning attacks and to understand its behaviors.
\end{itemize}

The rest of the paper is organized as follows. We review related work in Section 2. We define our problems in Section 3. We introduce the details of \ours in Section 4. Extensive experiments and their results are illustrated and analyzed in Section 5. We conclude the paper in Section 6.
\section{Related Work}
In this section, we briefly review related works, including graph neural networks, adversarial attack and defense on graphs.

\subsection{Graph Neural Networks} \label{related:gnn}
In general, graph neural networks refer to all deep learning methods for graph data \cite{wu2019comprehensive,DBLP:conf/kdd/0001WAT19,DBLP:conf/sdm/MaWAYT19,ma2019attacking,ding2019graph,ding2019deep}. It can be generally categorized into two categories, i.e., spectral-based and spatial-based. 
Spectral-based GNNs define  ``convolution'' following spectral graph theory~\cite{bruna2013spectral}. The first generation of GCNs are developed by \citeauthor{bruna2013spectral} \cite{bruna2013spectral} using spectral graph theory. Various spectral-based GCNs are developed later on \cite{defferrard2016convolutional, kipf2016semi, henaff2015deep, li2018adaptive}. 
To improve efficiency, spatial-based GNNs are  proposed to overcome this issue \cite{hamilton2017inductive,monti2017geometric,niepert2016learning,gao2018large}.
Because spatial-based GNNs directly aggregate neighbor nodes as the convolution, and are trained on mini-batches, they are more scalable than spectral-based ones.
Recently, \citeauthor{velivckovic2017graph} \cite{velivckovic2017graph} propose graph attention network (GAT) that leverages self-attention of neighbor nodes for the aggregation process. The major idea of GATs \cite{zhang2018gaan} is focusing on most important neighbors and assign higher weights to them during the information passing. 
However, \textit{existing GNNs aggregates neighbors' information for representation learning, making them vulnerable to adversarial attacks, especially perturbed edges added to the graph topology.} Next, we review adversarial attack and defense methods on graphs. 

\subsection{Adversarial Attack and Defense on Graphs}
Neural networks are widely criticized due to the lack of robustness \cite{goodfellow2014explaining,li2019nattack,li2018learning,cheng2018query,li2019regional,li2019click}, and the same to GNNs. Various adversarial attack methods have been designed, showing the vulnerability of GNNs \cite{dai2018adversarial,bojchevski2019adversarial,chen2018fast,zugner2019certifiable,xu2019adversarial}.
There are two major categories of adversarial attack methods, namely evasion attack and poisoning attack.
Evasion attack focuses on generating fake samples for a trained model. \citeauthor{dai2018adversarial} \cite{dai2018adversarial} introduce an evasion attack algorithm based on reinforcement learning. 
On the contrary, poisoning attack changes training data, which can decrease the performance of GNNs significantly. 
For example, \citeauthor{zugner2018adversarial} \cite{zugner2018adversarial} propose \textit{nettack} which make GNNs fail on any selected node by modifying its neighbor connections.
They further develop \textit{metattack} \cite{zugner2018adversarial2} that reduces the overall performance of GNNs. 
Comparing with evasion attack, poisoning attack methods are usually stronger and can lead to an extremely low performance \cite{zugner2018adversarial,zhu2019robust,sun2019node}, because of its destruction of training data. Besides, it is almost impossible to clean up a graph which is already poisoned. 
Therefore, we focus on defending the poisoning attack of graph data in this paper.

How to improve the robustness of GNNs against adversarial poising attacks is attracting increasing attention and initial efforts have been taken~\cite{xu2019topology,wu2019adversarial,zhu2019robust,jin2019power}.
For example, \citeauthor{wu2019adversarial} \cite{wu2019adversarial} utilize the Jaccard similarity of features to prune perturbed graphs with the assumption that connected nodes should have high feature similarity. RGCN in~\cite{zhu2019robust} adopts
Gaussian distributions as the node representations in each convolutional layer to absorb the effects of adversarial changes in the variances of the Gaussian distributions. 
The basic idea of aforementioned robust GNNs against poisoning attack is to alleviate the negative effects of the perturbed edges. However, perturbed edges are treated equally as normal edges during aggregation in existing robust GNNs.

The proposed \ours is inherently different from existing works: (\textbf{i}) instead of purely trained on the poisoned target graph, adopting clean graphs with similar domains to learn the ability of penalizing perturbations; and (\textbf{ii}) investigating meta-learning to transfer such ability to the target poisoned graph for improving the robustness. 
\section{Preliminaries}
\subsection{Notations}
We use $\G = (\V, \E, \X)$ to denote a graph, where $\V = \{v_1, \dots, v_N\}$ is the set of $N$ nodes, $\E \subseteq \V \times \V$ represents the set of edges, and $\X = \{\x_1, \dots, \x_N\}$ indicates node features. In a semi-supervised setting, partial nodes come with labels and are defined as $\V^L$, where the corresponding label for node $v$ is denoted by $y_v$. Note that the topology structure of $\G$ is damaged, and the original clean version is unknown.
In addition to the poisoned graph $\G$, we assume there exists $M$ clean graphs sharing similar domains with $\G$. For example, when $\G$ is the citation network of publications in data mining field, a similar graph can be another citation network from physics. We use $\{\bg_1, \dots, \bg_M\}$ to represent clean graphs. Similarly, each clean graph consists of nodes and edges. We use $\mathbf{V}_i^L$ to denote the labeled nodes in graph $\bg_i$.

\subsection{Basic GNN Design}
We introduce the general architecture of a graph neural network. A graph neural network contains multiple layers. Each layer transforms its input node features to another Euclidean space as output. Different from fully-connected layers, a GNN layer takes first-order neighbors' information into consideration when transforming the feature vector of a node. This ``message-passing'' mechanism ensures the initial features of any two nodes can affect each other even if they are faraway neighbors, along with the network going deeper.
The input node features to the $l$-th layer in an $L$-layer GNN can be represented by a set of vectors $\H^{l} = \{\h_1^{l}, \dots, \h_N^{l}\}, \h_i^{l} \in \R^{d_{l}}$, where $\h_i^l$ corresponds to $v_i$.
Obviously, $\H^1 = \X$.
The output node features of the $l$-th layer, which also formulate the input to the next layer, are generated as follows:
\begin{equation}
\label{eqn:general_layer}
    \h_i^{l+1} = \text{Update}\big [\h_i^{l}, \text{Agg}(\h_j^{l} | j\in\N_i)\big ]
\end{equation}
where $\N_i$ is the set of first-order neighbors of node $i$, $\text{Agg}(\cdot)$ indicates a generic aggregation function on neighbor nodes, and $\text{Update}(\cdot)$ is an update function that generates a new node representation vector from the previous one and messages from neighbors. Most graph neural networks follow the above definition. For example, \citeauthor{hamilton2017inductive} \cite{hamilton2017inductive} introduce mean, pooling and LSTM as the aggregation function, \citeauthor{velivckovic2017graph} \cite{velivckovic2017graph} leverage self-attention mechanism to update node representations.
A GNN can be represented by a parameterized function $f_\theta$ where $\theta$ represents parameters, the loss function can be represented as $\L_c(\theta)$.
In semi-supervised learning, the cross-entropy loss function for node classification takes the form:
\begin{equation}
\small\label{eqn:loss}
    \L_c(\theta) = -\sum_{v \in \V^L} y_v\log \hat{y}_v,
\end{equation}
where $\hat{y}_v$ is the predicted label generated by passing the output from the final GNN layer to a softmax function.

\subsection{Problem Definition}
The problem of exploring clean graphs for learning a robust GNN against poisoning attacks on a target graph is formally defined as:
\begin{mypro}
\textit{Given the target graph $\G$ that is poisoned with adversarial edges, a set of clean graphs $\{\bg_1, \dots, \bg_M\}$ from similar domain as $\G$, and the partially labeled nodes of each graph (i.e., $\{\mathbf{V}_1^L, \dots, \mathbf{V}_M^L; \V^L\}$), we aim at learning a robust GNN to predict the unlabeled nodes of $\mathcal{G}$.}
\end{mypro}
It is worth noting that, in this paper, we learn a robust GNN for semi-supervised node classification. The proposed \our is a general framework for learning robust GNN of various graph mining tasks such as link prediction.

\section{Proposed Framework}
\begin{figure}
    \centering
    \includegraphics[width=0.9\columnwidth]{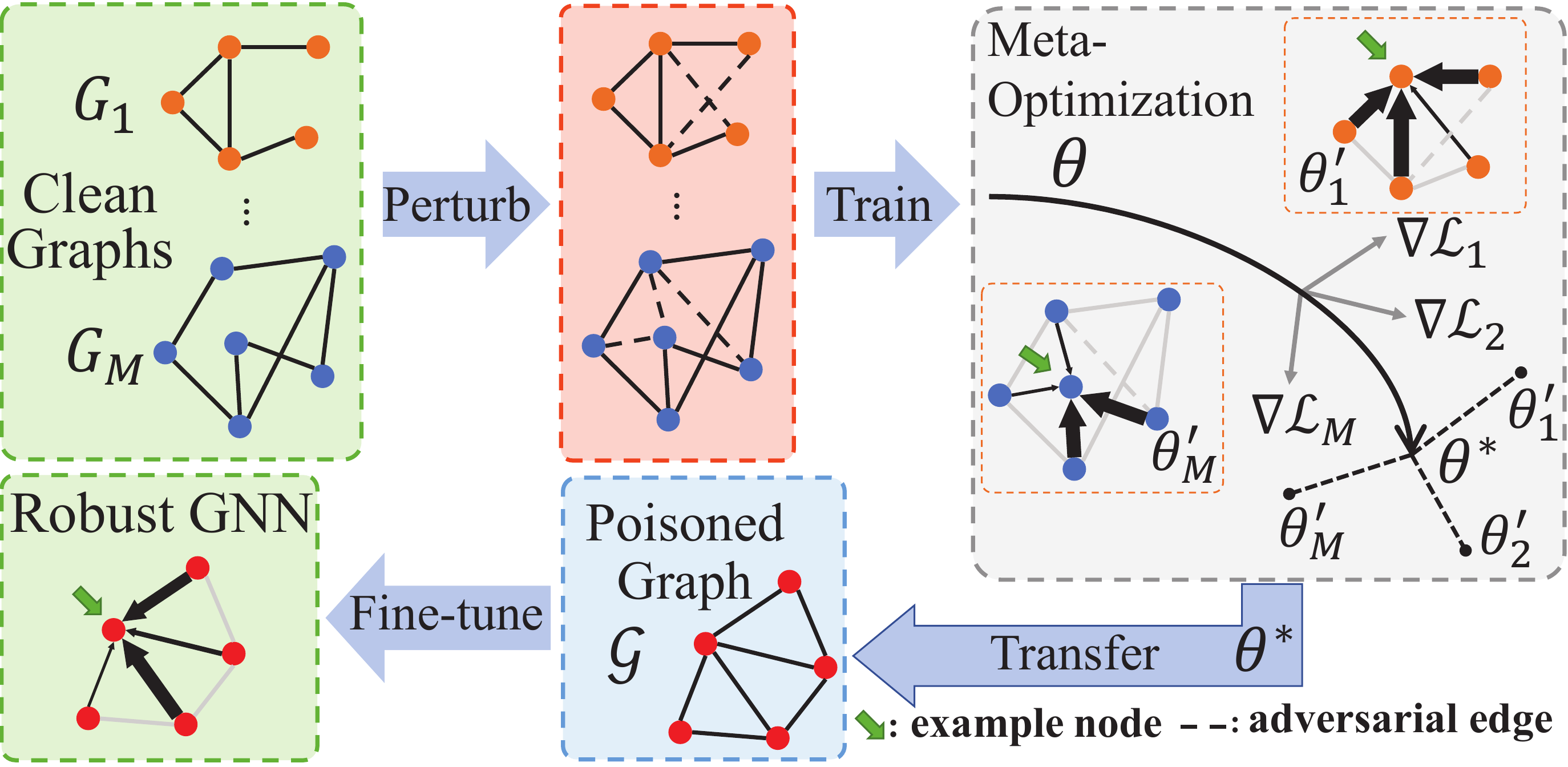}
    % \vspace{-1.2em}
    \caption{Overall framework of \ours. Thicker arrows indicate higher attention coefficients. $\theta^*$ denotes the model initialization from meta-optimization.}
    \label{fig:overall}
    \vspace{-2em}
\end{figure}
In this section, we give the details of \ours. An illustration of the framework is shown in Figure \ref{fig:overall}. Firstly, clean graphs $\{\bg_1, \dots, \bg_M\}$ are introduced to generate perturbed edges. The generated perturbations then serve as supervised knowledge to train a model initialization for \ours using meta-optimization. Finally, we fine-tune the initialization on the target poisoned graph for the best performance. Thanks to the meta-optimization, the ability to reduce negative effects of adversarial attack is retained after adapting to $\G$. In the following sections, we introduce technical details of \ours.

\subsection{Penalized Aggregation Mechanism} \label{method:attn_penalty}
We begin by analyzing the reason why GNNs are vulnerable to adversarial attacks with the general definition of GNNs in Equation \ref{eqn:general_layer}.
Suppose the graph data fed into a GNN is perturbed, the aggregation function $\text{Agg}(\cdot)$ treats ``fake'' neighbors equally as normal ones, and propagates their information to update other nodes. 
As a result, GNNs fail to generate desired outputs under influence of adversarial attacks.
Consequently, if messages passing through perturbed edges are filtered, the aggregation function will focus on ``true'' neighbors. In an ideal condition, GNNs can work well if all perturbed edges produced by attackers are ignored.

Motivated by above analysis, we design a novel GNN with penalized aggregation mechanism (\ours) which automatically restrict the message-passing through perturbed edge.
Firstly, we adopt similar implementation from \cite{vaswani2017attention} and define the self-attention coefficient $\text{a}_{ij}^l$ for node features of $v_i$ and $v_j$ on the $l$-the layer using a non-linear function:
\begin{equation}
\small\label{eqn:attn_coef}
    \text{a}_{ij}^l =
    \text{LeakyReLU} \big ( (\mathbf{a}^l)^\top [\W^l \h_i^l \oplus \W^l \h_j^l] \big ),
\end{equation}
where $\mathbf{a}^l$ and $\W^l$ are parameters, $\top$ represents the transposition, and $\oplus$ indicates the concatenation of vectors.
Note that coefficients are only defined for first-order neighbors. 
Take $v_i$ as an example,we only compute $\text{a}_{ij}^l$ for $j \in \N_i$, which is the set of direct neighbors of $v_i$. The attention coefficients related to $v_i$ are further normalized among all nodes in $\N_i$ for comparable scores:
\begin{equation}
\small
\label{eqn:attn_score}
    \alpha_{ij}^l = \frac{
    \text{exp}\big (\text{a}_{ij}^l \big )
    }{\sum_{k\in\N_i}
    \text{exp}\big (\text{a}_{ik}^l \big )
    }.
\end{equation}
We use normalized attention coefficient scores to generate a linear combination of their corresponding node features. The linear combination process serves as the aggregating process, and its results are utilized to update node features.
More concretely, a graph neural network layer is constructed as follows:
\begin{equation} 
\small\label{eqn:gat}
    \h_i^{l+1} = \sigma \big ( \sum_{j\in \N_j} \alpha_{ij}^{l}  \W^{l} \h_j^{l} \big ).
\end{equation}
A similar definition can be found in \cite{velivckovic2017graph}.
Clearly, the above design of GNN layer cannot discriminate perturbed edges, let alone alleviate their negative effects on the ``message-passing'' mechanism, because there is no supervision to teach it how to honor normal edges and punish perturbed ones.
A natural solution to this problem is reducing the attention coefficients for all perturbed edges in a poisoned graph. Noticing the exponential rectifier in Equation \ref{eqn:attn_score}, a lower attention coefficient only allows little information passing through its corresponding edge, which mitigate negative effects if the edge is an adversarial one. Moreover, since normalized attention coefficient scores of one node always sum up to 1, reducing the attention coefficient for perturbed edges will also introduce more attention to clean neighbors. To measure the attention coefficients received by perturbed edges, we propose the following metric:
\begin{equation} \label{eqn:attn_penalty} \small
    \mathcal{S}_p= \sum_{l=1}^{L} \sum_{e_{ij}\in\P} \text{a}_{ij}^l,
\end{equation}
where $L$ is the total number of layers in the network, and $\P$ denotes the perturbed edges. Generally, a smaller $\mathcal{S}_p$ indicates less attention coefficients received by adversarial edges. To further train GNNs such that a lower $\mathcal{S}_p$ is guaranteed, we design the following loss function to penalize perturbed edges:
\begin{equation} \label{eqn:attn_dist} \small
    \L_{dist} = - \min \Big ( \eta, \  \mathop{\mathbb{E}}_{\substack{e_{ij} \in \E \backslash \P \\ 1 \leq l \leq L}} \text{a}_{ij}^l - \mathop{\mathbb{E}}_{\substack{e_{ij} \in \P \\  1 \leq l \leq L}} \text{a}_{ij}^l \Big ),
\end{equation}
where $\eta$ is a hyper parameter controlling the margin between mean values of two distributions, $\E \backslash \P$ represents normal edges in the graph, and $\mathbb{E}$ computes the expectation. Using the expectation of attention coefficients for all normal edges as an anchor, $\L_{dist}$ aims at reducing the averaged attention coefficient of perturbed edges, until a certain discrepancy of $\eta$ between these two mean values is satisfied. 
Note that minimizing $\mathcal{S}_p$ directly instead of $\L_{dist}$ will lead to unstable attention coefficients, making \ours hard to converge. 
The expectations of attention coefficients are estimated by their empirical means:
\begin{align}
    \small \mathop{\mathbb{E}}_{\substack{e_{ij} \in \E \backslash \P \\ 1 \leq l \leq L}} \text{a}_{ij}^l & = \frac{1}{L|\E \backslash \P|} \sum_{l=1}^L  \sum_{e_{ij} \in \E \backslash \P} \text{a}_{ij}^l, \\
    \small \mathop{\mathbb{E}}_{\substack{e_{ij} \in \P \\  1 \leq l \leq L}} \text{a}_{ij}^l &  = \frac{1}{L|\P|} \sum_{l=1}^L \sum_{e_{ij} \in \P} \text{a}_{ij}^l,
\end{align}
where $|\cdot|$ denotes the cardinality of a set.
We combine $\L_{dist}$ with the original cross-entropy loss $\L_c$ and create the following learning objective for \ours:
\begin{equation} \label{eqn:objective} \small
   \min_{\theta} \L = \min_{\theta}(\L_c + \lambda \L_{dist}),
\end{equation}
where $\lambda$ balances the semi-supervised classification loss and the attention coefficient scores on perturbed edges.

Training \ours with the above objective directly is non-trivial, because it is unlikely to distinguish exact perturbed edges $\P$ from normal edges in a poisoned graph. However, it is practical to discover vulnerable edges from clean graphs with adversarial attack methods on graphs. 
For example, \textit{metattack} poisons a clean graph to reduces the performance of GNNs by adding adversarial edges, which can be treated as the set $\P$.
Therefore, we explore clean graphs from domains similar to the poisoned graph.
Specifically, as shown in Figure~\ref{fig:overall}, we first inject perturbation edges to clean graphs using adversarial attack methods, then leverage those adversarial counterparts to train the ability to penalize perturbed edges. 
Such ability is further transferred to GNNs on the target graph, so that the robustness is improved.
In the following section, we discuss how we transfer the ability to penalize perturbed edges from clean graphs to the target poisoned graph in detail.

\subsection{Transfer with Meta-Optimization} \label{meta_optim}
As discussed above, it is very challenging to train \ours for a poisoned graph because the adversarial edge distribution remains unknown.
We turn to exploit clean graphs from similar domains to create adversarial counterparts that serve as supervised knowledge.
One simple solution to utilize them is pre-training \ours on clean graphs with perturbations, which formulate the set of adversarial edges $\P$. Then the pre-trained model is fine-tuned on target graph $\G$ purely with the node classification objective.
However, the performance of pre-training with clean graphs and adversarial edges is rather limited, because graphs have different data distributions, making it difficult to equip GNNs with a generalized ability to discriminate perturbations. 
Our experimental results in Section \ref{ablation} also confirm the above analysis.

In recent years, meta-learning has shown promising results in various applications \cite{santoro2016meta,vinyals2016matching,yao2019learning,yao2019graph}. The goal of meta-learning is to train a model on a variety of learning tasks, such that it can solve new tasks with a small amount or even no supervision knowledge \cite{hochreiter2001learning,finn2017model,yao2019hierarchically}.
\citeauthor{finn2017model} \cite{finn2017model} propose model-agnostic meta-learning algorithm  where the model is trained explicitly such that a small number of gradient steps and few training data from a new task can also produce good generalization performance on that task. 
This motivates us to train a meta model with a generalized ability to penalize perturbed edges (i.e., assign lower attention coefficients).
The meta model serve as the initialization of \ours, and its fast-adaptation capability helps retain such penalizing ability as much as possible on the target poisoned graph. 
To achieve the goal, we propose a meta-optimization algorithm that trains the initialization of \ours. With manually generated perturbations on clean graphs, \ours receive full supervision and its initialization preserve the penalizing ability. 
Further fine-tuned model on the poisoned graph $\G$ is able to defend adversarial attacks and maintain an excellent performance.

We begin with generating perturbations on clean graphs. State-of-the-art adversarial attack method for graph -- \textit{metattack} \cite{zugner2018adversarial2} is chosen. Let $\P_i$ represent the set of adversarial edges created for clean graph $\bg_i$. 
Next, we define learning tasks for the meta-optimization. 
The learning objective of any task is defined in Equation \ref{eqn:objective}, which aims at classifying nodes accurately while assigning low attention coefficient scores to perturbed edges on its corresponding graph. Let $\T_i$ denote the specific task for $\bg_i$. Namely, there are $M$ tasks in accordance with clean graphs.
Because clean graphs are specified for every task, we use $\L_{\T_i}(\theta)$ to denote the loss function of task $\T_i$. 
We then compile support sets and query sets for learning tasks. Labeled nodes from each clean graph is split into two groups -- one for the support set and the other as the query set. Let $\S_i$ and $\Q_i$  denote the support set and the query set for $\bg_i$, respectively. 

Given $M$ learning tasks, the optimization algorithm first adapts the initial model parameters to every learning task separately. 
Formally, $\theta$ becomes $\theta^\prime_{i}$ when adapting to $\T_i$. We use gradient descent to compute the updated model parameter $\theta^\prime_{i}$. The gradient w.r.t $\theta^\prime_{i}$ is evaluated using  $\L_{\T_i}(\theta)$ on corresponding support set $\S_i$, and the initial model parameters $\theta$ are updated as follows:
\begin{equation} \label{eqn:maml_inner}
    \theta^\prime_{i} = \theta - \alpha \nabla_\theta \L_{\T_i}(\theta),
\end{equation}
where $\alpha$ controls the learning rate. 
Note that only one gradient step is shown in Equation \ref{eqn:maml_inner}, but using multiple gradient updates is a straightforward extension, as suggested by \cite{finn2017model}. 
There are $M$ different versions of the initial model (i.e., $f_{\theta_i^\prime},\cdots, f_{\theta_M^\prime}$) constructed in accordance with learning tasks.

The model parameters are trained by optimizing for the performance of $f_{\theta_i^\prime}$ with respect to $\theta$ across all tasks.
More concretely, we define the following objective function for the meta-optimization:
\begin{equation} \small
    \min_{\theta} \sum_{i=1}^M \L_{\T_i}({\theta^\prime_{i}})
    = \min_{\theta} \sum_{i=1}^M \L_{\T_i}({\theta - \alpha \nabla_\theta \L_{\T_i}(\theta)}).
\end{equation}
Because both classifying nodes and penalizing adversarial edges are considered by the objective of \ours, model parameters will preserve the ability to reduce the negative effects from adversarial attacks while maintaining a high accuracy for the classification.
Note that we perform meta-optimization over $\theta$ with the objective computed using the updated model parameters $\theta^\prime_{i}$ for all tasks.
Consequently, model parameters are optimized such that few numbers of gradient steps on a new task will produce maximally effective behavior on that task.
The characteristic of fast-adaptation on new tasks would help the model retain the ability to penalize perturbed edges on $\G$, which is proved by the experiential results in Section \ref{attn_vis}.
Formally, stochastic gradient descent (SGD) is used to update model parameters $\theta$ cross tasks:
\begin{align} \small
    \theta  & \leftarrow \theta - \beta \nabla_\theta \sum_{i=1}^{M} \L_{\T_i}({\theta_{i}^\prime}).
\end{align}
In practice, the above gradients are estimated using labeled nodes from query sets $\S_i$ of all tasks. Our empirical results suggest that splitting support sets and query sets on-the-fly through iterations of the meta-optimization improves overall performance. We adopt this strategy for the training procedure of \ours.

\vskip 0.5em
\noindent{}\textbf{Training Algorithm} An overview of the training procedure of \ours is illustrated in Algorithm \ref{alg}. 
% Adversarial edges are injected to clean graphs using \textit{metattack} (Line 2 to 4). 
% Support sets and query sets are randomly split on-the-fly for each task in Line 7.
% We then adapt initial model parameter $\theta$ to $\theta_i^\prime$ for each task $\T_i$. 
% The model parameters are updated by optimizing for the performance of all $f_{\theta_i^\prime}$ w.r.t $\theta$ using labeled nodes from query sets (Line 11).
% Finally, we adapt the trained model initialization to the targeted poisoned graph by minimizing the classification loss $\L_c$ on $\G$.

\begin{algorithm}
	\caption{The training framework of \ours}
    \label{alg}
    \KwIn{$\G$ and $\{\bg_1, \dots, \bg_M\}$}
    \KwOut{Model parameters $\theta$}
	Randomly initialize $\theta$\;
	\For{$\bg_i = \bg_1, \dots, \bg_M$}{
	    Select perturbed edge set $\P_i$ with \textit{metattack}\;
	}
	\While{not early-stop}{
	    \For{$\bg_i = \bg_1, \dots, \bg_M$}{
    	    Split labeled nodes of $\bg_i$ into support set $\S_i$ and $\Q_i$\;
    	   % $\theta_i^\prime \leftarrow \theta$\;
    	    Evaluating $\nabla_{\theta} \L_{\T_i}({\theta})$ with $\S_i$ and $\L_{\T_i}$\;
    	    Compute adapted parameters $\theta_i^\prime$ with gradient descent: $\theta^\prime_{i} \leftarrow \theta - \alpha \nabla_{\theta} \L_{\T_i}({\theta})$\;
	    }
	    Update $\theta$ on $\{\Q_1, \dots, \Q_M\}$ with: $\theta \leftarrow \theta - \beta \nabla_\theta \sum_{i=1}^{M} \L_{\T_i}({\theta_{i}^\prime})$\;
	}
	Fine-tune $\theta$ on $\G$\ use $\L_c$;
\end{algorithm}
\section{Experiments}
In this section, we conduct experiments to evaluate the effectiveness of \our. We aim to answer the following questions:
\begin{itemize}[leftmargin=*]
    \item Can \our outperform existing robust GNNs under representative and state-of-the-art adversarial attacks on graphs?
    \item How the penalized aggregation mechanism and the meta-optimization algorithm contribute to \our?
    \item How sensitive of \our on the hyper-parameters?
\end{itemize}
Next, we start by introducing the experimental settings followed by experiments on node classification to answer these questions.

\subsection{Experimental Setup}
\subsubsection{Datasets.}
To conduct comprehensive studies of \our, we conduct experiments under two different settings: 
\begin{itemize}[leftmargin=*]
    \item \textit{Same-domain setting}: We sample the poisoned graph and clean graphs from the same data distribution. Two popular benchmark networks (i.e., \textit{Pubmed} \cite{sen2008collective} and \textit{Reddit} \cite{hamilton2017inductive}) are selected as large graphs. 
    \textit{Pubmed} is a citation network where nodes are documents and edges represent citations; \textit{Reddit} is compiled from reddit.com where nodes are threads and edges denote two threads are commented by a same user. Both graphs build nodal features using averaged word embedding vectors \cite{pennington2014glove} of documents/threads.
    We create desired graphs using sub-graphs of the large graph. Each of them is randomly split into 5 similar-size non-overlapping sub-graphs. One graph is perturbed as the poisoned graph, while the remained ones are used as clean graphs. 
    
    \item \textit{Similar-domain setting}: We put \ours in real-world settings where graphs come from different scenarios. More concretely, we compile two datasets from Yelp Review\footnote{https://www.yelp.com/dataset}, which contains point-of-interests (POIs) and user reviews from various cities in Northern American. Firstly, each city in Yelp Review is transferred into a graph, where nodes are POIs, nodal features are averaged word-embedding vector \cite{pennington2014glove} of all reviews that a POI received, and binary labels are created to tell whether corresponding POIs are restaurants. We further define edges using co-reviews (i.e., reviews from the same author). 
    Graphs from different cities have different data distribution because of the differences in tastes, culture, lifestyle, etc.
    The first dataset (Yelp-Small) contains four middle-scale cities including Cleveland, Madison, Mississauga, and Glendale where Cleveland is perturbed as $\G$. The second dataset (Yelp-Large) contains top-3 largest cities including Charlotte, Phoenix, and Toronto. Specifically, we inject adversarial edges to the graph from Toronto to validate the transferability of \ours because Toronto is a foreign city compared with others.
\end{itemize}
We itemize statistics of datasets in Table \ref{tab:data}. We randomly select 10\% of nodes for training, 20\% for validation and remained for testing on all datasets (i.e., on $\G$). 40\% nodes from each clean graph are selected to build support and query sets, while remained ones are treated as unlabeled. Support sets and query sets are equally split on-the-fly randomly for each iteration of the meta-optimization (i.e., after $\theta$ is updated) to ensure the maximum performance.

\begin{table}[t]
    \small
    \centering
    \caption{Statistics of datasets}
    \vspace{-1.5em}
    \begin{tabular}{c|cccc}
    \hline \hline
         &  Pubmed & Reddit & Yelp-Small & Yelp-Large \\ \hline
      Avg. \# of nodes   & 1061 & 3180 & 3426 & 15757 \\
      Avg. \# of edges & 2614 & 14950 & 90431 & 160893 \\
      \# of features & 500 & 503 & 200 & 25 \\
      \# of classes & 3 & 7 & 2 & 2 \\ \hline
    \end{tabular}
    \vspace{-1.5em}
    \label{tab:data}
\end{table}
\subsubsection{Attack Methods.}
To evaluate how robust \our is under different attack methods and settings, three representative and state-of-the-art adversarial attack methods on graphs are chosen:
\begin{itemize}[leftmargin=*]
    
    \item Non-Targeted Attack: Non-targeted attack aims at reducing the overall performance of GNNs.
    We adopt \textit{metattack} \cite{zugner2018adversarial2} for non-targeted attack, which is also state-of-the-art adversarial attack method on graph data. We increase the perturbation rate (i.e., number of perturbed edges over all normal edges) from 0 to 30\%, by a step size of 5\% (10\% for Yelp-Large dataset due to the high computational cost of \textit{metattack}). We use the setting with best attack performance according to \cite{zugner2018adversarial2}.
    \item Targeted Attack: Targeted attack focuses on misclassifying specific target nodes. \textit{nettack} \cite{zugner2018adversarial} is adopted as the targeted attack method. 
    Specifically, we first randomly perturb 500 nodes with \textit{nettack} on target graph, then randomly assign them to training, validating, and testing sets according to their proportions (i.e., 1:2:7). This creates a realistic setting since not all nodes will be attacked (hacked) in a real-world scenario, and perturbations can happen in training, validating and testing sets. We adopt the original setting for \textit{nettack} from~\cite{zugner2018adversarial}.
    \item Random Attack: Random attack randomly select some node pairs, and flip their connectivity (i.e., remove existing edges and connect non-adjacent nodes). It can be treated as an injecting random noises to a clean graph.
    The ratio of the number of flipped edges to the number of clean edges varies from 0 to 100\% with a step size of 20\%.
\end{itemize}
We evaluate compared methods against state-of-the-art non-targeted attack method \textit{metattack} on all datasets. We analyze the performances against targeted attack on Reddit and Yelp-Large datasets. For random attack, we compare each method on Pubmed and Yelp-Small datasets as a complementary. Consistent results are observed on remained datasets.

\begin{table*}[!t]
\small
    \setlength{\tabcolsep}{8pt}
    \caption{Node classification performance (Accuracy$\pm$Std) under non-targeted \textit{metattack} \cite{zugner2018adversarial2}}\label{tab:nontarget}
    \vspace{-1.5em}
    \centering
    \begin{tabular}{c|c|ccccccc}
\hline \hline
              Dataset      & Ptb Rate (\%)        & 0              & 5              & 10             & 15             & 20             & 25             & 30             \\ \hline
\multirow{6}{*}{Pubmed}     & GCN                  & 77.81$\pm$0.34 & 76.00$\pm$0.24 & 74.74$\pm$0.55 & 73.69$\pm$0.37 & 70.39$\pm$0.32 & 68.78$\pm$0.56 & 67.13$\pm$0.32 \\
                            & GAT                  & 74.28$\pm$1.80 & 70.19$\pm$1.59 & 69.36$\pm$1.76 & 68.79$\pm$1.34 & 68.29$\pm$1.53 & 66.35$\pm$1.95 & 65.47$\pm$1.99 \\
                            & PreProcess           & 73.69$\pm$0.42 & 73.49$\pm$0.29 & 73.76$\pm$0.45 & 73.60$\pm$0.26 & 73.85$\pm$0.48 & 73.46$\pm$0.55 & 73.65$\pm$0.36 \\
                            & RGCN                 & 77.81$\pm$0.24 & 78.07$\pm$0.21 & 74.86$\pm$0.37 & 74.31$\pm$0.35 & 70.83$\pm$0.28 & 67.63$\pm$0.21 & 66.89$\pm$0.48 \\
                            & VPN              & 77.92$\pm$0.93 & 75.83$\pm$1.14 & 74.03$\pm$2.84 & 74.31$\pm$0.93 & 70.14$\pm$1.26 & 68.47$\pm$1.11 & 66.53$\pm$1.09 \\
                            & \our & \textbf{82.92$\pm$0.13} & \textbf{81.67$\pm$0.21} & \textbf{80.56$\pm$0.07} & \textbf{80.28$\pm$0.25} & \textbf{78.75$\pm$0.17} & \textbf{76.67$\pm$0.42} & \textbf{75.47$\pm$0.39} \\
                            \hline
\multirow{6}{*}{Reddit}     & GCN                  & \textbf{96.33$\pm$0.13} & 91.87$\pm$0.18 & 89.26$\pm$0.16 & 87.26$\pm$0.14 & 85.55$\pm$0.17 & 83.50$\pm$0.14 & 80.92$\pm$0.27 \\
                            & GAT                  & 93.81$\pm$0.35 & 92.13$\pm$0.49 & 89.88$\pm$0.60 & 87.91$\pm$0.45 & 85.43$\pm$0.61 & 83.40$\pm$0.39 & 81.27$\pm$0.38 \\
                            & PreProcess           & 95.22$\pm$0.18 & \textbf{95.14$\pm$0.19} & 88.40$\pm$0.35 & 87.00$\pm$0.27 & 85.70$\pm$0.25 & 83.59$\pm$0.27 & 81.17$\pm$0.30 \\
                            & RGCN                 & 93.15$\pm$0.44 & 89.20$\pm$0.37 & 85.81$\pm$0.35 & 83.58$\pm$0.29 & 81.83$\pm$0.42 & 80.22$\pm$0.36 & 76.42$\pm$0.82 \\
                            & VPN              & 95.91$\pm$0.17 & 91.95$\pm$0.17 & 89.03$\pm$0.28 & 86.97$\pm$0.15 & 85.38$\pm$0.24 & 83.49$\pm$0.29 & 80.85$\pm$0.28 \\
                            & \our & 95.80$\pm$0.11 & 94.35$\pm$0.33 & \textbf{92.16$\pm$0.49} & \textbf{90.74$\pm$0.56} & \textbf{88.44$\pm$0.20} & \textbf{86.60$\pm$0.17} & \textbf{84.45$\pm$0.34} \\
                            \hline
\multirow{6}{*}{Yelp-Small} & GCN                  & \textbf{87.27$\pm$0.31} & 74.54$\pm$0.98 & 73.44$\pm$0.35 & 73.30$\pm$0.83 & 72.16$\pm$0.88 & 69.70$\pm$0.90 & 68.55$\pm$0.85 \\
                            & GAT                  & 86.22$\pm$0.18 & 81.09$\pm$0.31 & 76.29$\pm$0.74 & 74.21$\pm$0.51 & 73.43$\pm$0.78 & 71.80$\pm$0.69 & 70.58$\pm$1.22 \\
                            & PreProcess           & 86.53$\pm$0.97 & 82.89$\pm$0.33 & 73.52$\pm$1.59 & 72.99$\pm$0.68 & 71.72$\pm$0.99 & 70.38$\pm$0.62 & 69.31$\pm$1.32 \\
                            & RGCN                 & 88.19$\pm$0.31 & 79.70$\pm$0.69 & 77.25$\pm$2.12 & 75.85$\pm$1.31 & 75.65$\pm$0.33 & 74.71$\pm$0.21 & 73.30$\pm$2.95 \\
                            & VPN              & 86.05$\pm$1.60 & 78.13$\pm$0.38 & 74.36$\pm$1.54 & 74.33$\pm$0.59 & 72.54$\pm$0.35 & 71.86$\pm$0.78 & 70.13$\pm$1.72 \\
                            & \our & 86.53$\pm$0.18 & \textbf{86.34$\pm$0.18} & \textbf{84.17$\pm$0.17} & \textbf{82.41$\pm$0.46} & \textbf{77.69$\pm$0.25} & \textbf{76.77$\pm$0.60} & \textbf{76.20$\pm$0.39} \\
                            \hline
\multirow{6}{*}{Yelp-Large} & GCN                  & 84.21$\pm$0.48 & $-$            & 80.96$\pm$1.66 & $-$            & 80.56$\pm$1.69 & $-$            & 78.64$\pm$0.46 \\
                            & GAT                  & 84.73$\pm$0.22 & $-$            & 81.25$\pm$0.36 & $-$            & 79.82$\pm$0.42 & $-$            & 77.81$\pm$0.39 \\
                            & PreProcess           & 84.54$\pm$0.25 & $-$            & 82.16$\pm$4.12 & $-$            & 78.80$\pm$2.17 & $-$            & 78.05$\pm$2.63 \\
                            & RGCN                 & \textbf{85.09$\pm$0.13} & $-$            & 79.42$\pm$0.27 & $-$            & 78.31$\pm$0.08 & $-$            & 77.74$\pm$0.12 \\
                            & VPN              & 84.36$\pm$0.23 & $-$            & 82.77$\pm$0.25 & $-$            & 80.64$\pm$2.41 & $-$            & 79.22$\pm$2.32 \\
                            & \our & 84.98$\pm$0.16 & $-$            & \textbf{84.66$\pm$0.09} & $-$            & \textbf{82.71$\pm$0.29} & $-$            & \textbf{81.48$\pm$0.12} \\
                            \hline
\end{tabular}
    % \vspace{-1em}
\end{table*}

\subsubsection{Baselines.}
We compare \our with representative and state-of-the-art GNNs and robust GNNs. The details are:
\begin{itemize}[leftmargin=*]
    \item \textbf{GCN} \cite{kipf2016semi}: GCN is a widely used graph neural network. It defines graph convolution via spectral analysis.  We adopt the most popular version from \cite{kipf2016semi}.
    \item \textbf{GAT} \cite{hamilton2017inductive}: As introduced in Section \ref{related:gnn}, GAT leverages multi-head self-attention to assign different weights to neighborhoods.
    \item \textbf{PreProcess} \cite{wu2019adversarial}: This method improves the robustness of GNNs by removing existing edges whose connected nodes have low feature similarities. Jaccard similarity is used sparse features and Cosine similarity is adopted for dense features.
    %Only sparse features are considered in the original paper. We extend the solution to graphs with dense features, using cosine similarity instead of Jaccard similarity for features.
    \item \textbf{RGCN} \cite{zhu2019robust}: RGCN aims to defend against adversarial edges with Gaussian distributions as the latent node representation in hidden layers to absorb the negative effects of adversarial edges.
    \item \textbf{VPN} \cite{jin2019power}: Different from GCN, parameters of VPN are trained on a family of powered graphs of $\G$. The family of powered graphs increases the spatial field of normal graph convolution, thus improves the robustness.
\end{itemize}
\textit{Note that PreProcess, RGCN and VPN are state-of-the-art robust GNNs developed to defend against adversarial attacks on graphs.}

\subsubsection{Settings and Parameters.} \label{sec:settings_parameters}
% We use semi-supervised node classification as the evaluation task. For each perturbation rate, we randomly select 10\% of nodes from $\G$ as labeled nodes for training, 20\% of nodes for validation, and the remaining 70\% nodes as unlabeled nodes for testing. Same split is used for all methods for a fair comparison. 
We report the averaged results of 10 runs for all experiments. We deploy a multi-head mechanism \cite{vaswani2017attention} to enhance the performance of self-attention. We adopt \textit{metattack} to generate perturbations on clean graphs.
% regardless of {\color{blue}as well as?} the attack method on target graph.
% \suhang{add support set and query set for clean graphs}
% We adopt the original implementations for all baselines. 
All hyper-parameters are tuned on the validation set to achieve the best performance. For a fair comparison, following a common way~\cite{zhu2019robust}, we fix the number of layers to 2 and the total number of hidden units per layer to 64 for all compared models. 
We set $\lambda$ to 1.0 and $\eta$ to 100 for all settings. Parameter sensitivity on $\lambda$ and $\eta$ will be analyzed in Section~\ref{sec:parameter_sensitivity}. We perform  5 gradient steps to estimate $\theta^\prime$ as suggested by \cite{finn2017model}.

\subsection{Robustness Comparison}

To answer the first question, we evaluate the robustness of \our  under various adversarial attack scenarios with comparison to baseline methods. We adopt semi-supervised node classification as our evaluation task as described in Section~\ref{sec:settings_parameters}.

\subsubsection{Defense Against Non-Targeted Attack.}
We first conduct experiments under non-targeted attack on four datasets. Each experiment is conducted 10 times. The average accuracy with standard deviation is reported in Table \ref{tab:nontarget}. From the table, we make the following observations: (i) As illustrated, the accuracy of vanilla GCN and GAT decays rapidly when the perturbation rate goes higher, while other robust GNN models achieve relatively higher performance in most cases. This suggests the necessity of improving the robustness of GNN models; (ii) The prepossessing-based method shows consistent results on the Pubmed dataset with sparse features. However, it fails for other datasets. Because the feature similarity and neighbor relationship are often complementary, purely relying on feature similarity to determining perturbation edges is not a promising solution. On the contrary, \our aims at learning the ability to detect and penalizing perturbations from data, which is more dynamic and reliable;
(iii) Comparing with RGCN, \our achieves higher performance under different scenarios. This is because \our successfully leverages clean graphs for improving the robustness. Moreover, instead of constraining model parameters with Gaussian distributions, \our directly restricts the attention coefficients of perturbed edges, which is more straightforward. The above observations articulate the efficacy of \ours, which successfully learns to penalize perturbations thanks to the meta-optimization on clean graphs. Lastly, we point out that \ours achieves slightly higher or comparable performance even if $\G$ is clean (i.e., no adversarial edges), showing the advantage of the meta-optimization process.

% \suhang{TODO: 1. outperform all and explain; 2. has comparable or slightly higher results even when there are no adversarial edges}
% Moreover, \our out-perform VPN, too. Alghouth the graph power \cite{jin2019power} improves the robustness of GNNs, its capability of resisting adversarial attacks is rather limited, which motivates us to design robust GNN models specifically for adversarial attacks.
% However, \our can consistently out-perform other baselines on all datasets and all perturbation rates. 
% This confirm the effectiveness of the proposed penalizing aggregation mechanism w.r.t eliminate the negative influence of perturbations.
% Lastly, \our show relatively higher performances on Pubmed and Reddit datasets. The potential reason is when clean graphs are more similar to $\G$, they contribute more to \our with the meta-optimization. 

\subsubsection{Defense Against Targeted Attack}
We further study how robust \our is under targeted attack.
% We select Reddit and Yelp-Large datasets to verify the robustness of each method under targeted attack.
% We evaluate \ours on Reddit and Yelp-Large datasets, as we have similar observations on the other two datasets. 
As shown in Table \ref{tab:target}, \our outperforms all the compared methods under targeted attack, with approximate 5\% performance improvements on both datasets compared with second accurate methods.
This confirms the reliability of \our against targeted attack.
Moreover, note that the perturbations of clean graphs are generated by \textit{metattack}, which is a non-target adversarial attack algorithm. We conclude that \ours does not rely on specific adversarial attack algorithm to train model initialization. The ability to penalize perturbation can be generalized to defend other adversarial attacks. A similar conclusion can be drawn from following experiments against random attack.

\begin{table*}[]
\small
\setlength{\tabcolsep}{8pt}
\renewcommand{\arraystretch}{1}
\centering
    \caption{Node classification accuracy under targeted attack.}
    \label{tab:target}
    \vskip -1.5em
\begin{tabular}{c|cccccc}
\hline \hline
Dataset                       & GCN            & GAT            & PreProcess     & RGCN           & VPN        & \our \\ \hline
% \multirow{2}{*}{Reddit}     & 95.77$\pm$0.11 & 91.04$\pm$0.23 & 95.27$\pm$0.09 & 90.67$\pm$0.15 & 96.01$\pm$0.14 & \textbf{96.64$\pm$0.17}       \\
                       Reddit      & 74.25$\pm$0.20 & 73.83$\pm$0.12 & 73.02$\pm$0.18 & 74.75$\pm$0.15 & 74.00$\pm$0.07 & \textbf{79.57$\pm$0.13}       \\ \hline
% \multirow{2}{*}{Yelp-Large} & W/o Ptb & 84.47$\pm$0.24  & 	\textbf{86.29$\pm$0.36} & 	85.07$\pm$0.61 & 	85.68$\pm$0.24 & 	84.22$\pm$0.24 & 	84.71$\pm$0.00  \\
                       Yelp-Large &   71.97$\pm$0.12 & 	71.12$\pm$0.73 & 	74.83$\pm$0.12	 & 77.01$\pm$0.24 & 	72.09$\pm$0.73 & 	\textbf{82.28$\pm$0.49}                     \\ \hline
\end{tabular}
    \vspace{-1.0em}
\end{table*}

% \vspace{-5pt}
\subsubsection{Defense Against Random Attack.}
\begin{figure}
    \centering
    \begin{subfigure}[b]{.22\textwidth}
      \centering
      \includegraphics[width=\columnwidth]{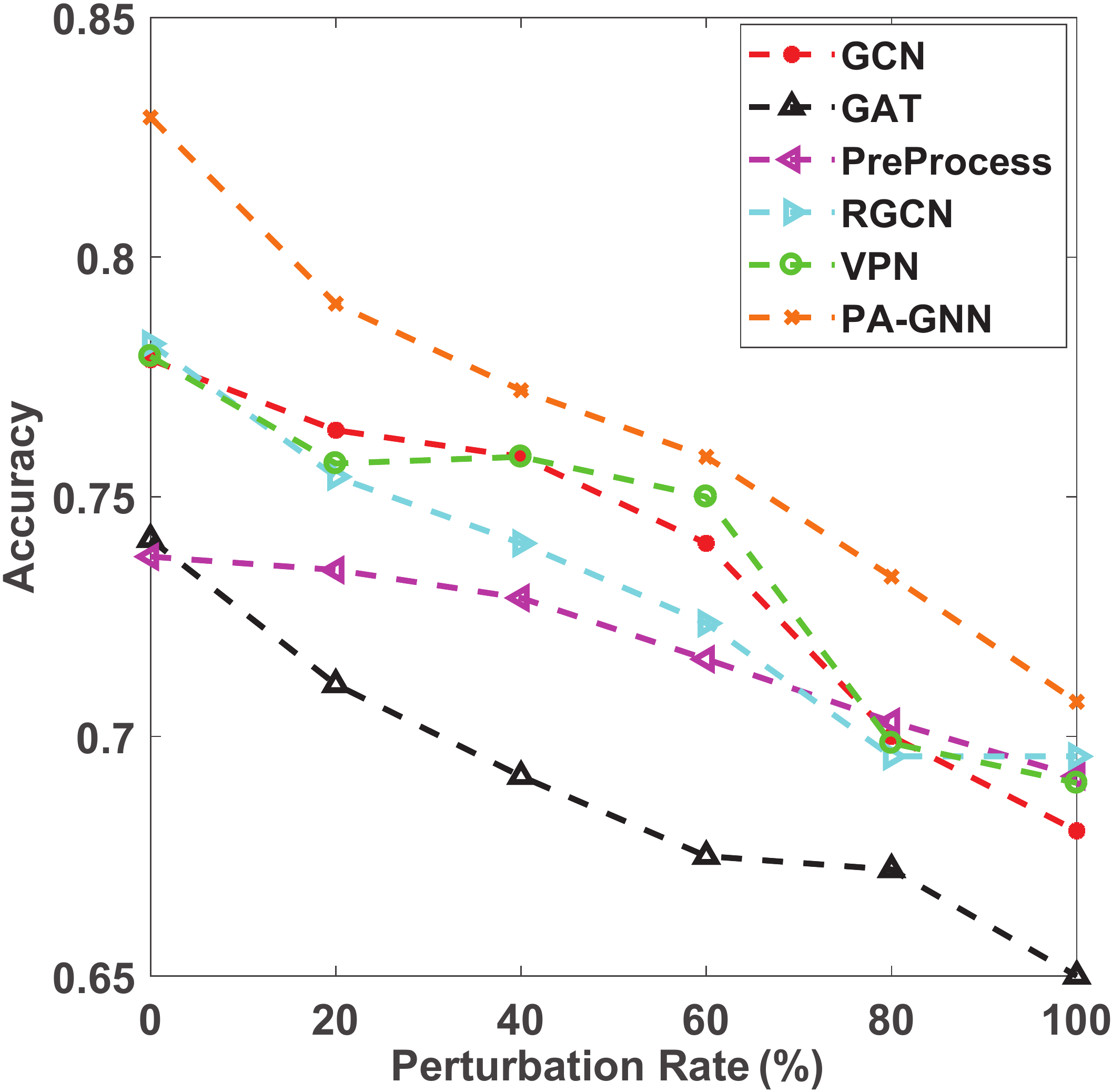}
      \vskip -0.5em
      \caption{Pubmed}
    %   \label{fig:attn_penalty_wot}
    \end{subfigure}
    \begin{subfigure}[b]{.22\textwidth}
      \centering
      \includegraphics[width=\columnwidth]{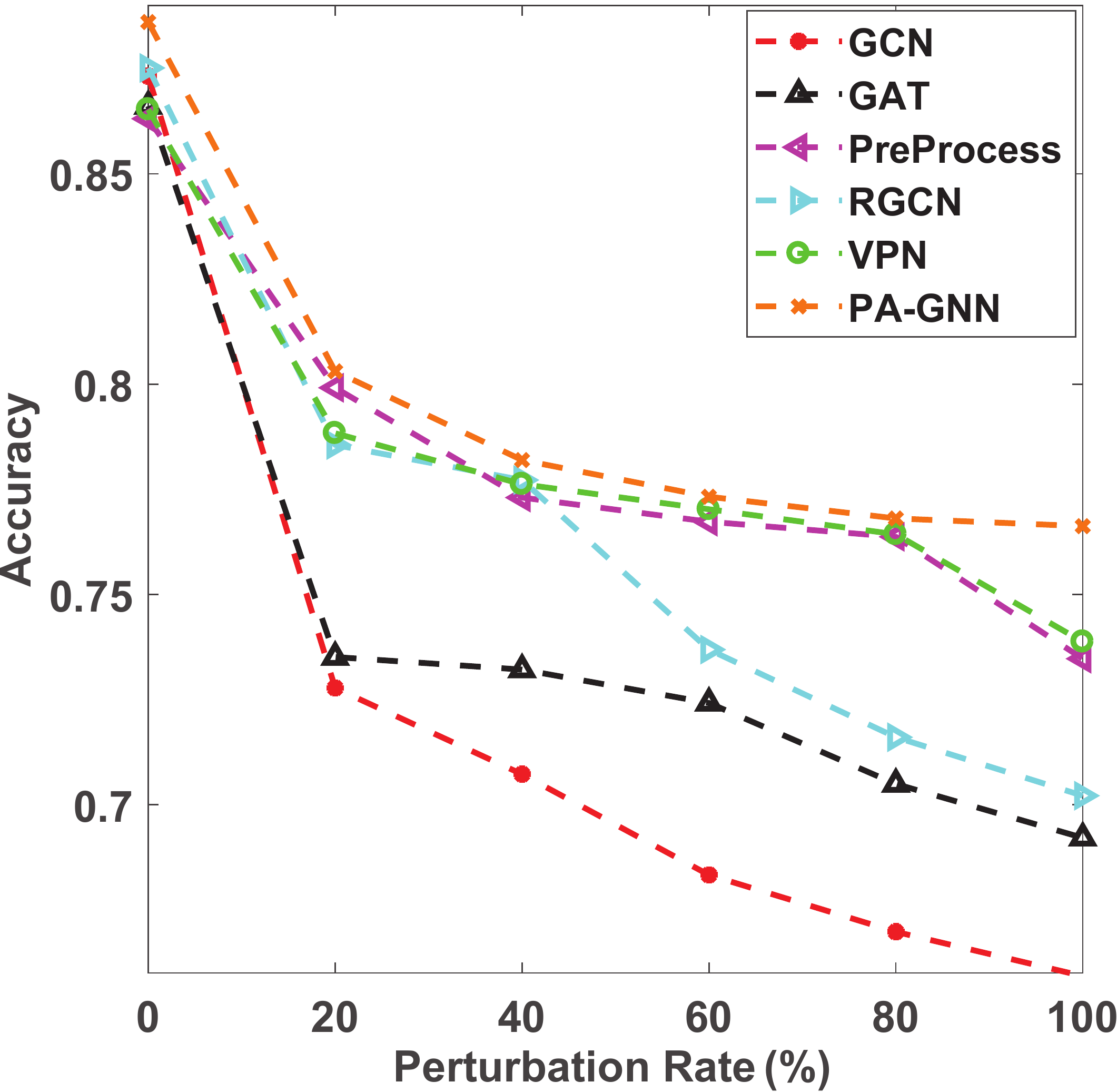}
      \vskip -0.5em
      \caption{Yelp-Small}
    %   \label{fig:attn_penalty_with}
    \end{subfigure}
    \vskip -1em
    \caption{Node classification accuracy under random attack.} 
    % \suhang{replace PowerUp}}
    \label{fig:random}
    \vskip -1.7em
\end{figure}

Finally, we evaluate all compared methods against random attack. As shown in Figure \ref{fig:random}, \ours consistently out-performs all compared methods. Thanks to the meta-optimization process, \ours successfully learns to penalize perturbations, and transfers such ability to target graph with a different kind of perturbation. Besides, the low performance of GAT indicates the vulnerability of the self-attention, which confirms the effectiveness of the proposed penalizing aggregation mechanism.

\begin{table*}[t]
\small
    \setlength{\tabcolsep}{8pt}
    \centering
    \caption{Node classification accuracy of ablations.}
    \vskip -1.5em
    \label{tab:ablation}
    \begin{tabular}{c|ccccccc}
    \hline \hline
        Ptb Rate (\%) & 0            & 5            & 10           & 15           & 20           & 25           & 30           \\ \hline
         $\text{\our}_{np}$  &	95.25$\pm$0.81  &	92.17$\pm$0.23  &	90.45$\pm$0.72  &	88.72$\pm$0.61  &	86.66$\pm$0.18  &	84.68$\pm$0.52  &	81.53$\pm$0.34 \\ 
         $\text{\our}_{2nd}$ & 77.11$\pm$0.67 &		75.43$\pm$1.11 &		71.18$\pm$1.24 &		68.51$\pm$1.95 &		64.86$\pm$1.59 &		63.16$\pm$1.29 &		61.08$\pm$1.07 \\
         $\text{\our}_{ft}$ & \textbf{96.72$\pm$0.09}	 &		91.89$\pm$0.14	 &		89.79$\pm$0.24	 &		87.56$\pm$0.25	 &		85.41$\pm$0.17	 &		83.88$\pm$0.35	 &		82.14$\pm$0.38 \\
         $\text{\our}_{jt}$ & 96.63$\pm$0.18	 &	92.13$\pm$0.19 &		88.62$\pm$0.35 &		87.00$\pm$0.27 &		84.65$\pm$0.25 &		82.75$\pm$0.27 &		81.20$\pm$0.30 \\\hline
         \our &  95.80$\pm$0.11 & \textbf{94.35$\pm$0.33} & \textbf{92.16$\pm$0.49} & \textbf{90.74$\pm$0.56} & \textbf{88.44$\pm$0.20} & \textbf{86.60$\pm$0.17} & \textbf{84.45$\pm$0.34} \\
\hline 
    \end{tabular}
    \vspace{-1em}
\end{table*}

\begin{table}[t]
\small
    \setlength{\tabcolsep}{8pt}
    \renewcommand{\arraystretch}{1.05}
    \centering
    \caption{Mean values of attention coefficients.}
    \label{tab:attn_dist}
    \vskip -1.5em
    \begin{tabular}{c|cc}
    \hline \hline
         & Normal edges  & Ptb. edges \\ \hline
        W/o penalty & 12.63 & 12.80\\ 
       With penalty  & 4.76 & 3.86\\ 
        \hline
    \end{tabular}
    \vspace{-1em}
\end{table}

% \vspace{-3pt}
\subsection{Ablation Study} \label{ablation}
To answer the second question, we conduct ablation studies to understand the penalized aggregation and meta-optimization algorithm. 

\subsubsection{Varying the Penalized Aggregation Mechanism.}
\label{attn_vis}
\begin{figure}[t]
    \centering
    \begin{subfigure}[b]{.23\textwidth}
      \centering
      \includegraphics[width=\columnwidth]{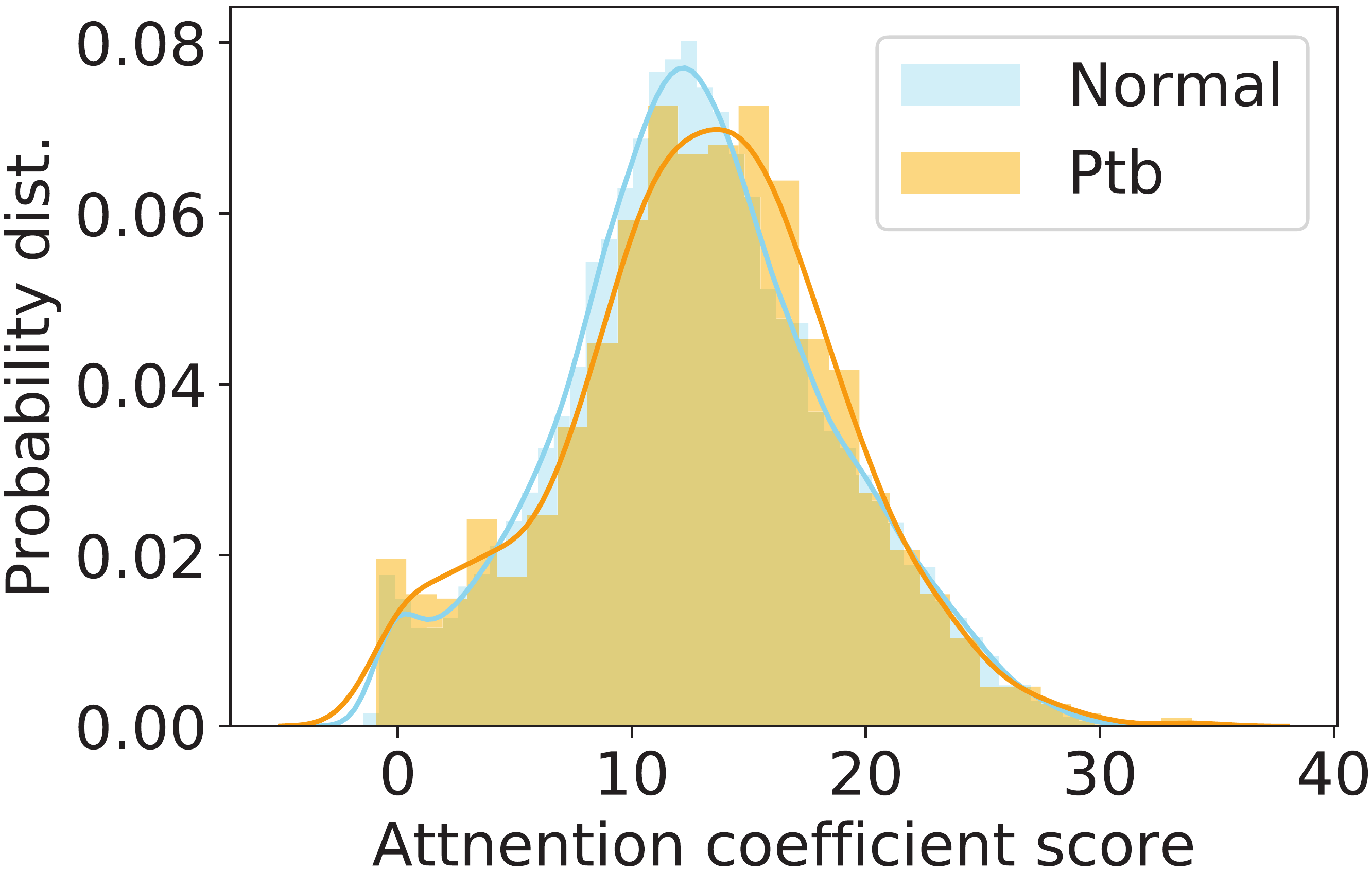}
      \vskip -0.5em
      \caption{W/o penalized aggregation.}
      \label{fig:attn_penalty_wot}
    \end{subfigure}
    \begin{subfigure}[b]{.23\textwidth}
      \centering
      \includegraphics[width=\columnwidth]{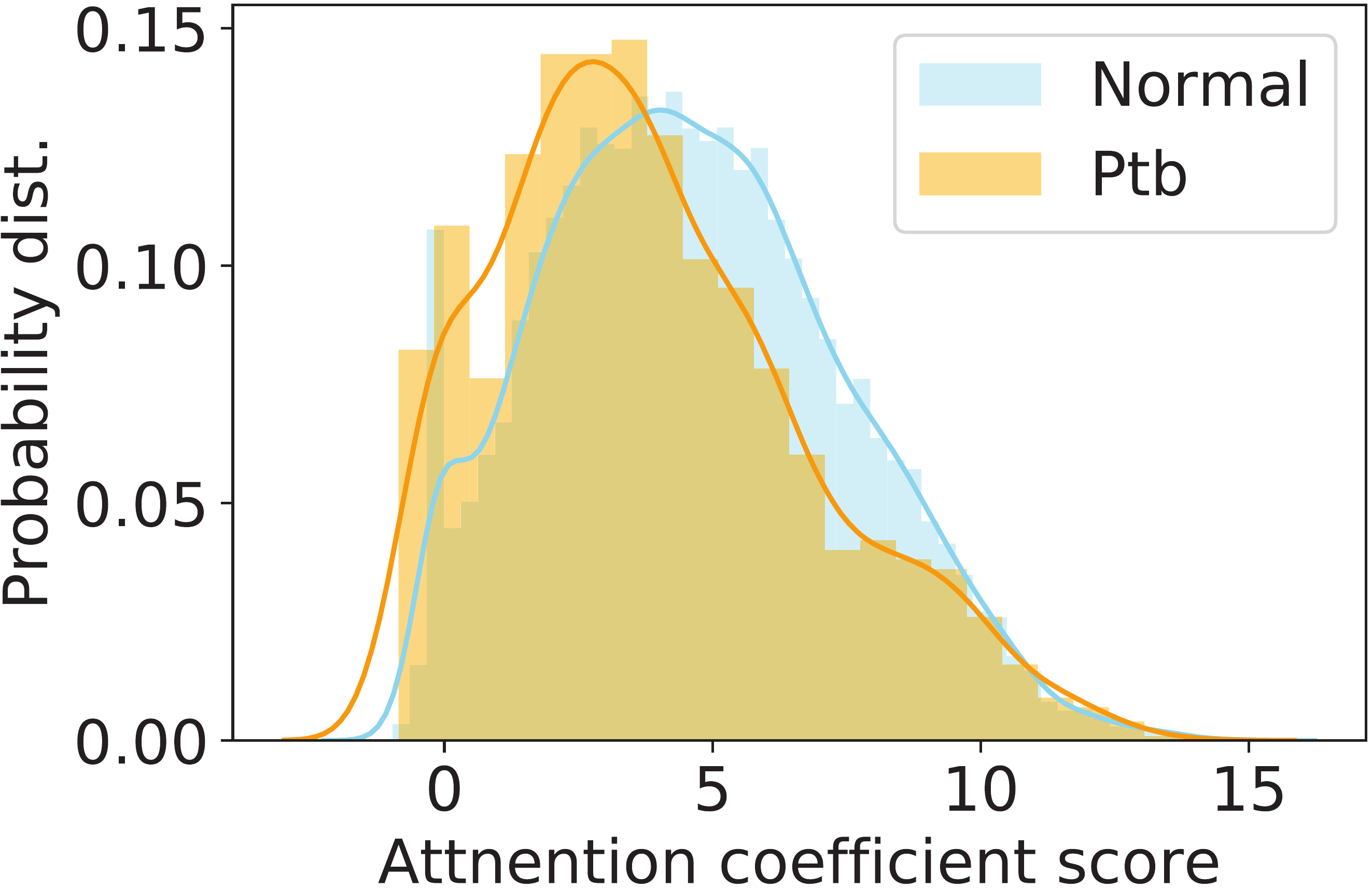}
      \vskip -0.5em
      \caption{With penalized aggregation.}
      \label{fig:attn_penalty_with}
    \end{subfigure}
    \vskip -1em
    \caption{Distributions of attention coefficients in \ours.}
    \label{fig:attn_dist}
    \vspace{-1.7em}
\end{figure}

We analyze the effect of proposed penalized aggregation mechanism from two aspects. 
Firstly, we propose $\text{\our}_{np}$, a variant of \ours that removes the penalized aggregation mechanism by setting $\lambda=0$. We validate $\text{\our}_{np}$ on Reddit dataset, and its performance against different perturbation rates is reported in Table \ref{tab:ablation}. As we can see, \our consistently out-performs $\text{\our}_{np}$ by 2\% of accuracy. The penalized aggregation mechanism limits negative effects from perturbed edges, in turns improves the performance on the target graph.
Secondly, we explore distributions of attention coefficient on the poisoned graph of \ours with/without the penalized aggregation mechanism. 
% We explore the fine-tuned \ours on the target poisoned graph. 
Specifically, the normalized distributions of attention coefficients for normal and perturbed edges are plotted in Figure \ref{fig:attn_dist}. 
We further report their mean values in Table \ref{tab:attn_dist}.
Without the penalized aggregation, perturbed edges obtain relatively higher attention coefficients. This explains how adversarial attacks hurt the aggregation process of a GNN.
As shown in Figure \ref{fig:attn_penalty_with}, normal edges receive relative higher attention coefficients through \ours, confirming the ability to penalize perturbations is transferable since \ours is fine-tuned merely with the node classification objective.
These observations reaffirm the effectiveness of the penalized aggregation mechanism and the meta-optimization algorithm, which successfully transfers the ability to penalize perturbations in the poisoned graph.
\subsubsection{Varying the Meta-Optimization Algorithm.}
\label{ablation:meta}
Next, we study the contribution of the meta-optimization algorithm. As discussed in Section \ref{meta_optim}, three ablations are created accordingly: $\text{\our}_{2nd}$, $\text{\our}_{ft}$, and $\text{\our}_{jt}$.
$\text{\our}_{2nd}$ ignores clean graphs and rely on a second-time attack to generate perturbed edges.
$\text{\our}_{ft}$ omit the meta-optimization process, training the model initialization on clean graphs and their adversarial counterparts jointly. We then fine-tune the initialization for $\G$ using the classification loss $\L_c$.
$\text{\our}_{jt}$ further simplifies $\text{\our}_{ft}$ by adding $\G$ to the joint training step. Note that we remove $\L_{dist}$ for $\G$ because detailed perturbation information is unknown for a poisoned graph.
All three variants are evaluated on Reddit dataset, and their performance is reported in Table \ref{tab:ablation}.

$\text{\our}_{2nd}$ performs the worst among all variations. Because perturbed edges from the adversarial attack can significantly hurt the accuracy, treating them as clean edges  is not a feasible solution.
$\text{\our}_{ft}$, and $\text{\our}_{jt}$ slightly out-perform \ours when $\G$ is clean. This is not amazing since more training data can contribute to the model. However, their performance decreases rapidly as the perturbation rate raises up. Because the data distribution of a perturbed graph is changed, barely aggregate all available data is not an optimal solution for defending adversarial attack. It is vital to design \ours which leverages clean graphs from similar domains for improving the robustness of GNNs.
At last, $\text{\our}_{np}$ consistently out-performs $\text{\our}_{ft}$, and $\text{\our}_{jt}$ in perturbed cases. shown advantages of the meta-optimization algorithm which utilizes clean graphs to train the model regardless of the penalized aggregation mechanism.

\subsection{Parameter Sensitivity Analysis} \label{sec:parameter_sensitivity}
\begin{figure}[h]
    \centering
    % \vskip -1em
    \begin{subfigure}{.2\textwidth}
      \centering
      \includegraphics[width=\columnwidth]{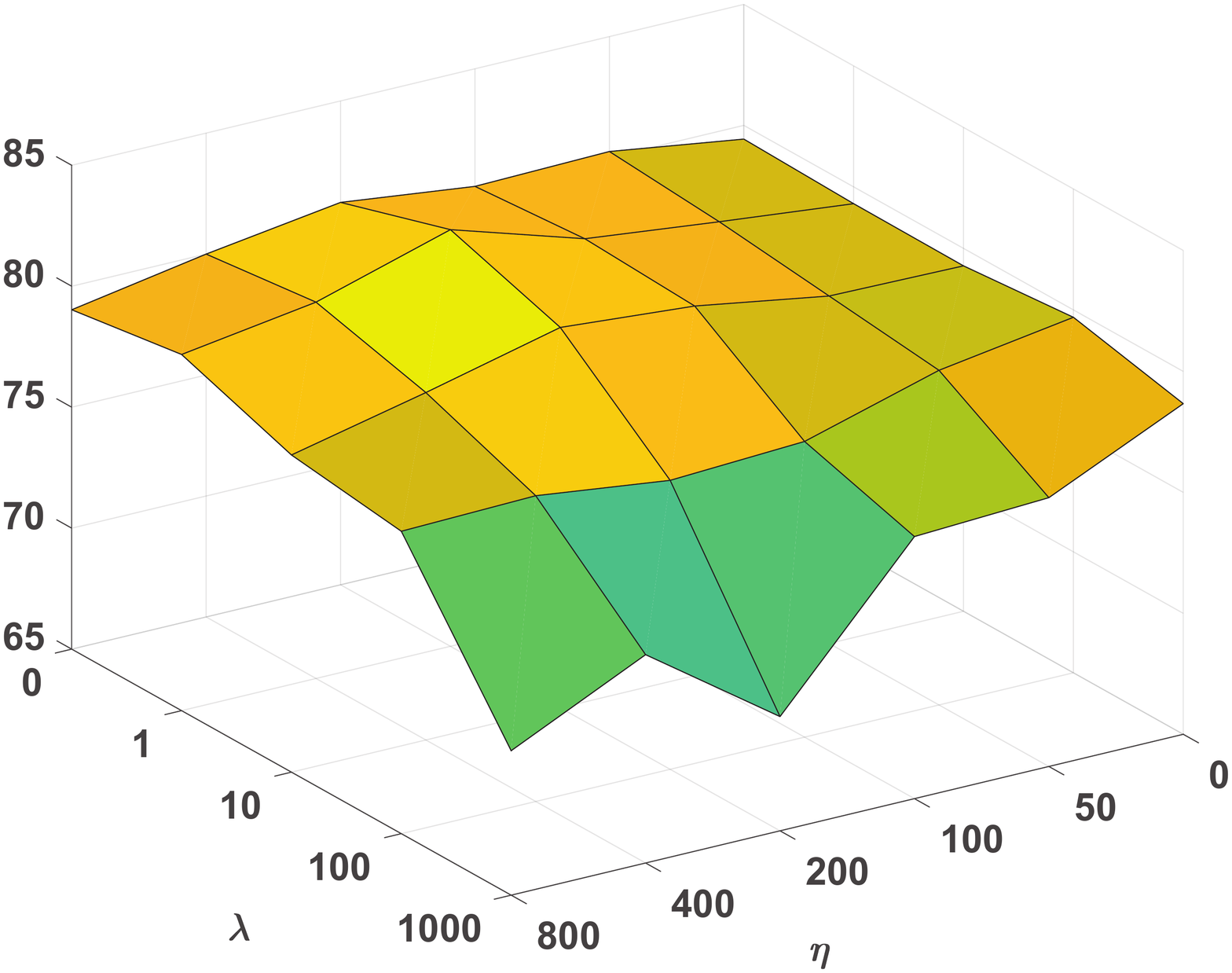}
    %   \vskip -0.5em
     \caption{Pubmed with 10\% Ptb.}
      \label{fig:para_pubmed}
    \end{subfigure} \quad
    \begin{subfigure}{.2\textwidth}
      \centering
      \includegraphics[width=\columnwidth]{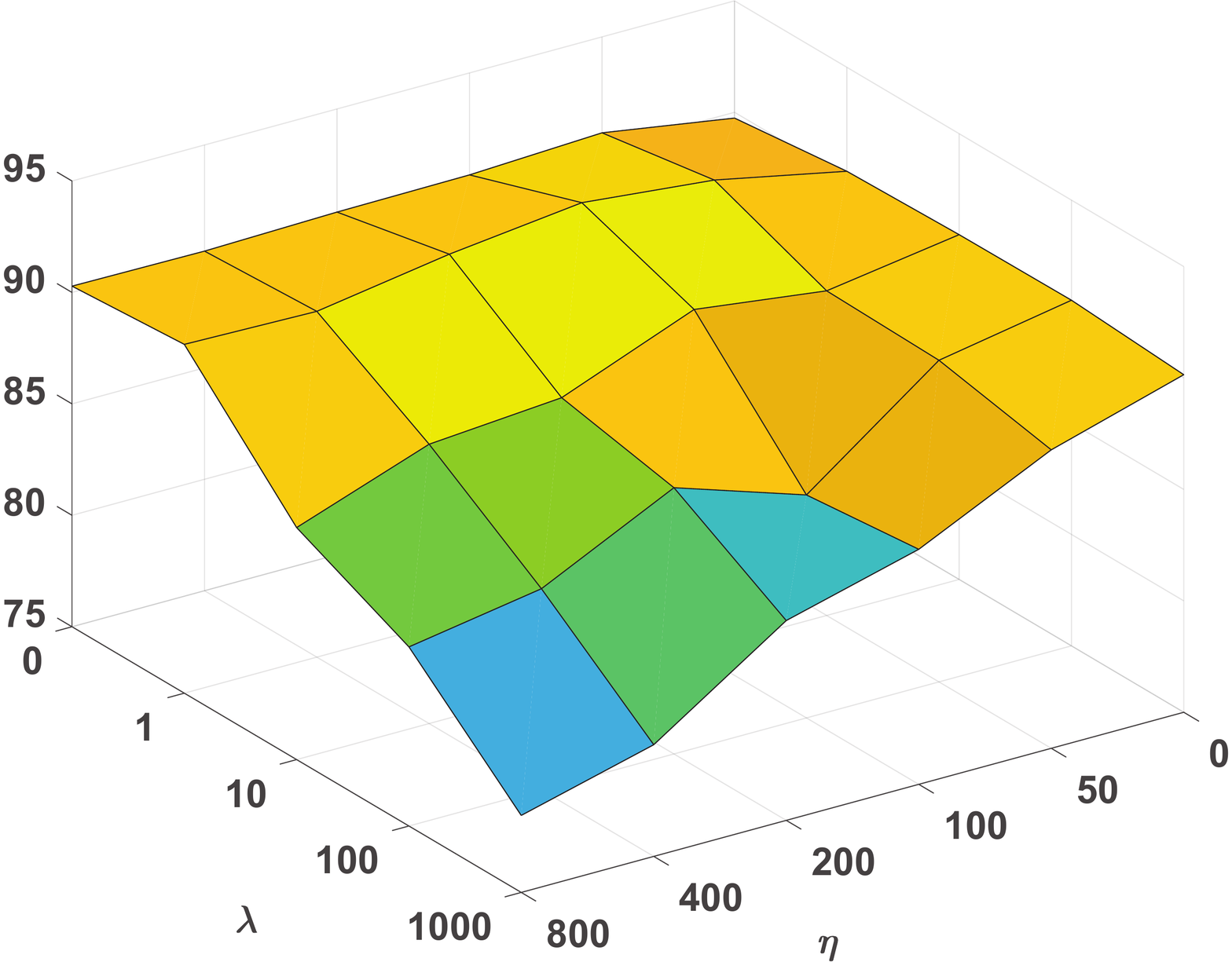}
    %   \vskip -0.5em
      \caption{Reddit with 10\% Ptb.}
      \label{fig:para_reddit}
    \end{subfigure}
    % \vskip -1.5em
    \caption{Parameter sensitivity analysis.}
    \label{fig:para}
    % \vskip -2em
\end{figure}

We investigate the sensitivity of $\eta$ and $\lambda$ for \ours. $\eta$ controls the penalty of perturbed edges, while $\lambda$ balances the classification objective and the penalized aggregation mechanism.
Generally, a larger $\eta$ pull the distribution of perturbed edges farther away from that of normal edges.
We explore the sensitivity on Pubmed and Reddit datasets, both with a 10\% perturbation rate. We alter $\eta$ and $\lambda$ among $\{0, 1, 10, 100, 1000\}$ and $\{0, 50, 100, 200, 400, 800\}$, respectively. The performance of \ours is illustrated in Figure \ref{fig:para}. As we can see, the accuracy of \ours is relatively smooth when parameters are within certain ranges. However, extremely large values of $\eta$ and $\lambda$ result in low performances on both datasets, which should be avoided in practice. Moreover, increasing $\lambda$ from 0 to 1 improves the accuracy on both datasets, demonstrating the proposed penalized aggregation mechanism can improve the robustness of \ours.

\section{Conclusion and Future Work}
In this paper, we study a new problem of exploring extra clean graphs for learning a robust GNN against the poisoning attacks on a target graph. We propose a new framework \our, that leverages penalized attention mechanism to learn the ability to reduce the negative impact from perturbations on clean graphs and meta-optimization to transfer the alleviation ability to the target poisoned graph. Experimental results of node classification tasks demonstrate the efficacy of \ours against different poisoning attacks. In the future, we would like to explore the potential of transfer learning for improving robustness on other models, such as community detection and graph classification.

\begin{acks}
This material is based upon work supported by, or in part by, the National Science Foundation (NSF) under grant \#1909702.
\end{acks}

\bibliographystyle{ACM-Reference-Format}
\bibliography{ref}

\end{document}